\documentclass[letterpaper, 10 pt, conference]{ieeeconf}  

\IEEEoverridecommandlockouts                              

\usepackage{amsmath,amssymb,amsfonts}
\usepackage{algorithmic}
\usepackage{graphicx}
\usepackage{textcomp}
\usepackage{xcolor}
\usepackage[export]{adjustbox}
\usepackage{cite}
\def\BibTeX{{\rm B\kern-.05em{\sc i\kern-.025em b}\kern-.08em
		T\kern-.1667em\lower.7ex\hbox{E}\kern-.125emX}}

\usepackage[linesnumbered,lined,boxed,commentsnumbered]{algorithm2e}
\usepackage{multicol}
 
\usepackage{float}
\usepackage[caption = false]{subfig}

\title{\LARGE \bf
 VIR-SLAM: Visual, Inertial, and Ranging SLAM\\ for single and multi-robot systems
}

\author{Yanjun Cao and Giovanni Beltrame
\thanks{M. Cao and Prof. Beltrame are with the Department of Computer and Software Engineering, \'Ecole Polytechnique de Montr\'eal, 2900 Boul \'Edouard-Montpetit, Qu\'ebec	CA \protect
		E-mail: (\emph{name.surname}@polymtl.ca).}%
}

\begin{document}

\maketitle
\thispagestyle{empty}
\pagestyle{empty}


\begin{abstract}
  Monocular cameras coupled with inertial measurements generally give high
  performance visual inertial odometry. However, drift can be significant with
  long trajectories, especially when the environment is visually
  challenging. In this paper, we propose a system that leverages
  ultra-wideband ranging with one static anchor placed in the environment to
  correct the accumulated error whenever the anchor is visible. We also use
  this setup for collaborative SLAM: different robots use mutual ranging (when
  available) and the common anchor to estimate the transformation between each
  other, facilitating map fusion.  Our system consists of two modules: a
  double layer ranging, visual, and inertial odometry for single robots, and a
  transformation estimation module for collaborative SLAM. We test our system
  on public datasets by simulating an ultra-wideband sensor as well as on real
  robots. Experiments show our method can outperform state-of-the-art
  visual-inertial odometry by more than 20\%. For visually challenging
  environments, our method works even the visual-inertial odometry has
  significant drift.  Furthermore, we can compute the collaborative SLAM
  transformation matrix at almost no extra computation cost.
\end{abstract}


\section{INTRODUCTION}


Robot localization is a fundamental topic in any mobile robot
application. Recent advances in robot hardware and software have boosted the
opportunity and demand for multi-robot systems for their inherent benefits,
such as high efficiency and robustness.

With a monocular camera and low-cost inertial measurement units, Visual
Inertial Odometry (VIO) is accepted as the minimal sensor configuration for
single robot state estimation and navigation considering size, wight, power and cost \cite{delmerico_benchmark_2018}.
Recent technical advances
\cite{lupton_preintegration_2012, forster_manifold_2017, qin_vinsmono_2018}
make VIO more and more robust and stable in many conditions. However, the
drift caused by accumulated error is still hard to control without loop
closures. Although loop closure is a natural part for SLAM system, the
requirements to close the loop rely much on the trajectory and
environment. For example, generating high-quality closures requires revisiting
the same location with a similar viewpoint. Furthermore, perception outliers
caused by illumination, self-similar environments, etc. are challenging for
loop closure.


In this paper, we use single extra sensor, a static ultra-wideband (UWB)
anchor, to improve the performance of robot localization. UWB technology has
attracted a lot of attention recent years for its accurate ranging performance
and long-distance support.  For example, the latest Apple IPhone at the time
of writing is equipped with a UWB chip (actually, it includes all sensors
needed in this paper).  Most available UWB systems
(e.g.~\cite{noauthor_loco_nodate}) use several (at least four for 3D and
three for 2D) calibrated anchors as a Global Positioning System (GPS) for
specific areas. This type of infrastructure is not applicable for the
exploration in unknown environments, which is one of the primary objectives of
SLAM. Therefore, we design our system to rely only on one anchor, which can be
dropped off at any moment by a robot during its mission. Our experiments show
this one anchor can improve the localization accuracy significantly.


\begin{figure}[h]
\begin{center}
\includegraphics[trim=0cm 0.2cm 1.2cm 0.3cm,clip, width=0.37\textwidth,center]{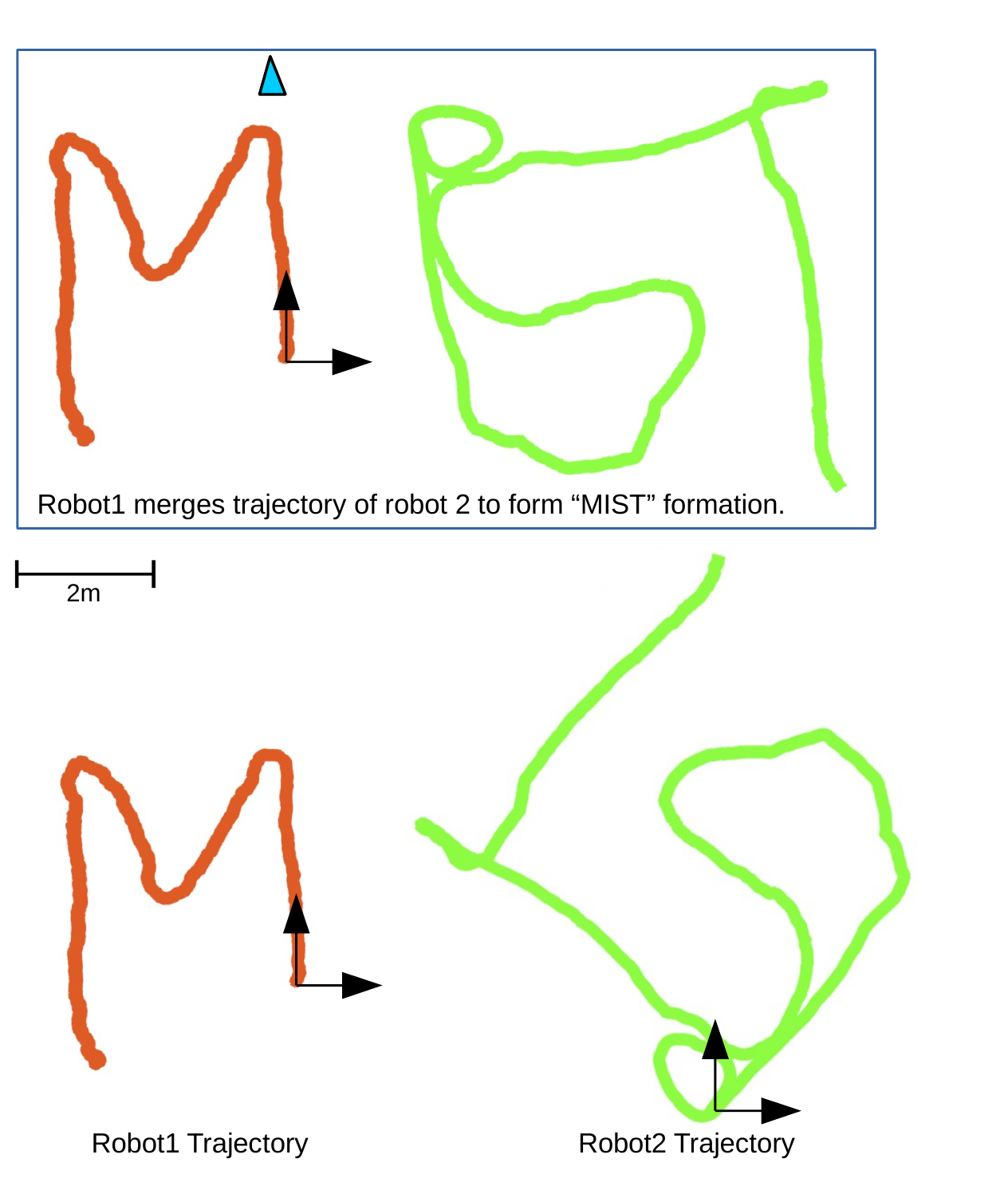}
\caption{ Two robots are manually controlled independently. 
  The first robot with
  a Realsense T265 and an UWB module draws an "M"  in its own
  coordinate frame. The other robot, carrying a Realsense D435, a UWB module and
  a Pixracer flight controller, draws "IST" in its frame.
  One static UWB anchor is placed in the environment (the blue triangle shown in Robot1 frame). Their 
  trajectories are shown at the bottom.  With two inter-robot communications and measurements,
  robot 1 can estimate the transformation matrix from robot 2, and then map robot
  2's trajectory in its own frame as the top picture shows, forming "MIST"
  (our lab).}
\label{fig:multirobotExp} 
\end{center}
\end{figure}

This setup has another significant benefit for multi-robot SLAM. Multi-robot
SLAM is a promising field, but much more challenging. One issue is the
estimation of the transformation matrix between
robots~\cite{saeedi_multi_2016}. Most available works rely on common features
to extract the relative pose of robots, which is a type of inter-robot loop
closure, which leads to similar constraints as single robot loop closures.  An
additional challenge for inter-robot loop closure is the need to exchange the
information required for loop closure among all robots, which can be
significant.

In our system, we can solve this challenge using the UWB sensor. When any two
robots move into their respective UWB ranging radius, they can exchange their
mutual ranging and anchor information 
and they can estimate their respective coordinate systems
transformations. Having the transformation matrix, a robot can correctly
project all the information from its neighbors onto its own frame. This
solution greatly simplifies multi-robot SLAM, minimizing the need for
inter-robot loop closure, which requires the exchange of feature databases,
the identification of loop closures and distributed pose graph optimization.

The last but not least reason for choosing UWB is its inherent advantage in
data association. When a range measurement is received, the identity of the
sender is recognized without extra effort, which can be of great assistance
for multi-robot systems. Moreover, UWB can provide low bandwidth communication
while ranging. For example, the state estimation of poses can be carried by
UWB packets while performing ranging. In the future, we believe a
monocular camera, inertial measurement units (IMU), and UWB will be a standard
minimal configuration for multi-robot systems.

\section{RELATED WORK}


\subsubsection{Single Robot SLAM with Visual, Inertial and Ranging Measurements}

SLAM has been the subject of intense research for more than thirty
years~\cite{cadena_slam_2016}. Monocular
visual-inertial odometry is a popular choice as it provides good state
estimation performance with a minimal sensor configuration. Although
state-of-the-art VIO algorithms (e.g. SVO \cite{forster_manifold_2017},
VINS-Mono~\cite{qin_vinsmono_2018}, DSO~\cite{ engel_direct_2018}) can reach
very high accuracy in relative translation and orientation, the accumulated
drift can still be an issue: any small orientation error can lead to large
end-point error. Our system leverages UWB ranging measurements to correct the
accumulated error. We developed our system based on
VINS-Mono~\cite{qin_vinsmono_2018}, which is a robust and versatile state
estimator which uses a sliding window tightly-coupled nonlinear optimization
for visual and IMU measurements.

UWB technology, as a localization solution on its own, has attracted a lot of
attention in recent years both in research and industry for its decimeter
localization accuracy. However, most results are based on a well-calibrated
multi-anchor setup~\cite{prorok_accurate_2014,mueller_fusing_2015,
  fang_graph_2018, tiemann_enhanced_2018}, which is not applicable for
navigation in unexplored, unstructured environments.  Wang et
al.~\cite{wang_ultra-wideband_2017} propose a system using camera, IMU and UWB
to bypass the complexity of loop closure. However, they still use multiple
pre-configured UWB anchors. Their UWB module provides coarse drift-free global
position and VIO identifies the local trajectory. On the contrary, in this
paper we use only one anchor, placed in an unspecified location.  Shi et
al.\cite{shi_visual-uwb_nodate} have a similar spirit to ours. They
start with one UWB anchor and keep dropping anchors from a moving
robot. Unfortunately, their experimental results are available only in
simulation for one sequence of the EuRoC dataset~\cite{burri_euroc_2016}, with
five simulated anchors. In our work, we focus on the use of a single anchor
setup. We design a double layer sliding window algorithm, which effectively
fuses accurate VIO and range constraints along the trajectory.

\subsubsection{Multi-Robot SLAM}

Multi-robot SLAM has gained recent attention for the increased viability and
accessibility of multi-robot systems. Saeedi et
al.\cite{saeedi_multiple-robot_2016} give a comprehensive review of
multi-robot SLAM and points out one key issue: relative pose estimation. Most
current multi-robot SLAM systems solve this issue by analyzing inter-robot
loop closures, either in centralized~\cite{schmuck_ccm-slam_2019} or
distributed~\cite{choudhary_distributed_2017, mangelson_pairwise_2018}
fashion. The distributed approach is more robust, but it is harder to
implement in practice: robots need to exchange map data to get the feature
database for future loop closures, and distributed optimization
usually requires additional communication and computation.

Ranging measurements can assist in relative pose estimation. Trawny et
al.\cite{trawny_3d_nodate} provide theoretical proofs and simulations that
show how six range measurements can be used to get the transformation matrix
between two robots. Martel et al.\cite{martel_unique_2019} extend the work and
adapt it to 4DOF relative pose estimation with a UWB setup, and use it for
merging maps for VR applications. Both methods use range measurements over
long trajectories. In our solution, robots can estimate the transformation
matrix as soon as they can get two measurements from their neighbors, which
meets the requirement of real-time transformation estimation during robot
rendezvous. These two methods~\cite{trawny_3d_nodate, martel_unique_2019}
represent a good solution when no common anchor is present and can be combined
with inter-robot loop closures to improve the transformation results.

In practice, most works in literature have explored the combination of
multiple UWB anchors with vision and/or IMU. In this paper, we focus on the
use of a single anchor in an arbitrary location, which is trivial to deploy as
a beacon in real exploration tasks. We propose a double layer sliding window
technique to combine VIO with UWB ranging, which produces drift-free state
estimation by leveraging VIO for its accurate short time relative pose
estimation, and range constraints for longer trajectories. Moreover, recorded
anchor ranges can help robots find the transformation matrix
efficiently with only two range measurements in multi-robot settings. Our
contributions can be summarized as:
\begin{itemize}
\item an UWB-aided SLAM system for single robots which outperforms state-of-the-art VIO;
\item a double layer sliding window algorithm that combined relative pose
  estimation from VIO and UWB ranging constrains;
\item an efficient method to estimate the transformation matrix between multiple robots.
\end{itemize}

\section{SYSTEM DESIGN}

In this section, we explain the preliminaries and symbols used in this
paper. We then detail the formulation of our ranging-aided visual inertial
SLAM, focusing on the cost factors of the optimization problem. Finally, we
introduce our solution for the estimation of the transformation matrix between
robots.

\begin{figure}
\begin{center}
\includegraphics[width=0.8\columnwidth]{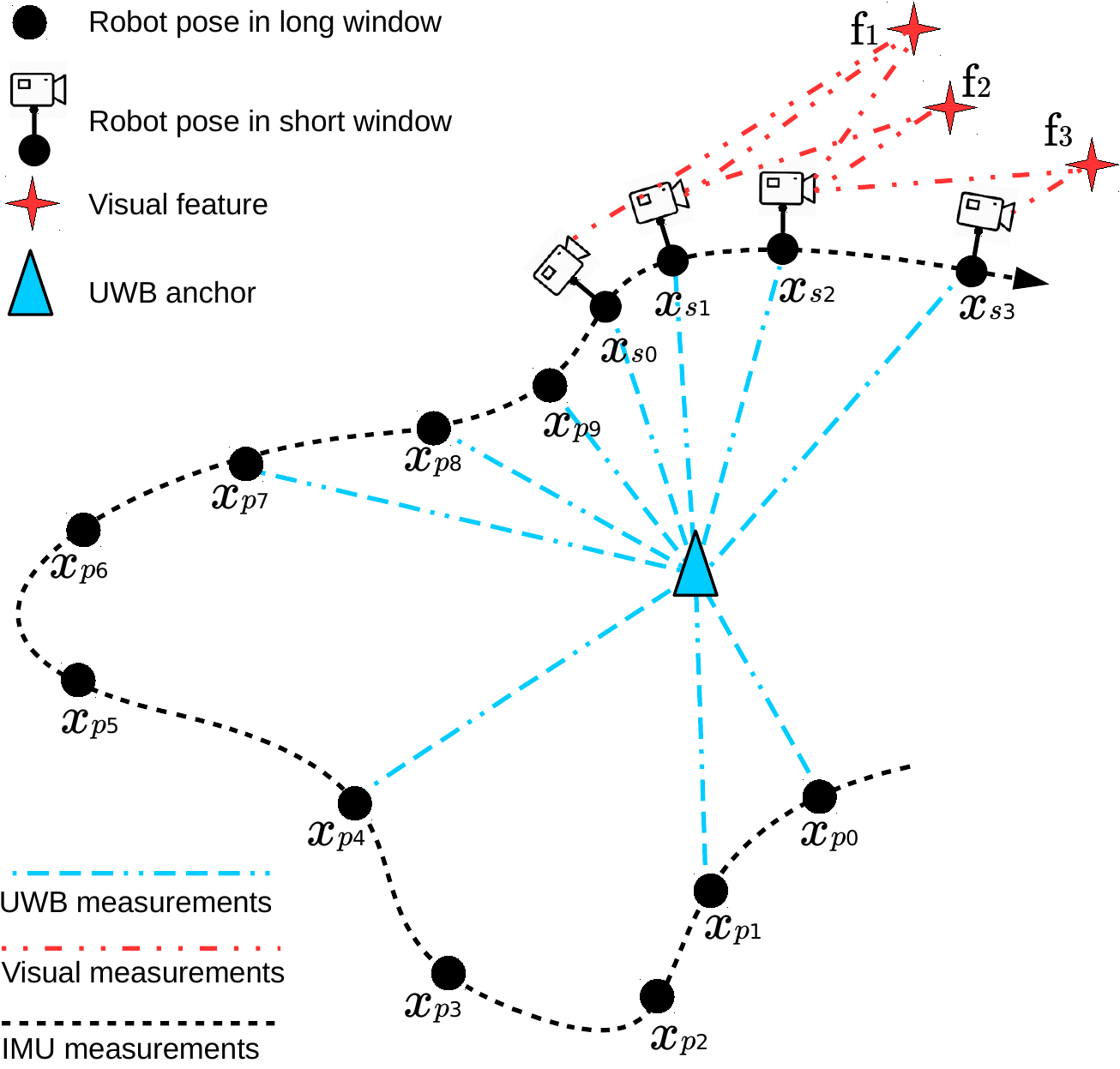}
\caption{Illustration of the scenario of VIR-SLAM, with a single anchor
  setup. Measurements from sensors are shown with different lines. Robot poses
  with the camera and IMU measurements belong to the short sliding window. All
  poses associated with UWB ranging to the anchor are kept for the long
  sliding window.}
\label{fig:problem} 
\end{center}
\end{figure}

\subsection{Single Robot Estimation Preliminaries}

We assume a robot carries three kinds of sensors (a monocular camera, an IMU
and UWB module), and moves in a 3D environment, as shown in
Fig.~\ref{fig:problem}. An UWB anchor is placed in the environment with an
unknown initial position. We define the world frame of the robot $i$ as
$[\;\;]^{{iw}}$, which is usually aligned with the first camera frame when the
robot starts its mission. The position of the anchor is expressed in the robot
world frame, denoted by $\mathbf{P}_A^{iw}$. We use $[\;\;]^{ib}$ to indicate
the body frame of robot $i$, and similarly $[\;\;]^{ic}$ for the camera
frame. Note that we do not define the UWB sensor frame because it is a scalar
measurement. The UWB ranging measurement is transferred to body frame by
considering the 3D offset of the UWB antenna in the body frame.


Classical VIO proposes an optimization formulation of states over
a sliding window with size $n$ as:
\begin{align}
\mathcal{X} &= [\mathbf{x}_0,\mathbf{x}_1,\dots,\mathbf{x}_n,l_0, l_1, \dots, l_m] \label{equ:classicalState}\\
\mathbf{x}_k &= [
\mathbf{p}_{b_k}^{iw}, \mathbf{v}_{b_k}^{iw}, \mathbf{q}_{b_k}^{iw}, \mathbf{b}_{a_k}^{ib}, \mathbf{b}_{w_k}^{ib}] \;\;\; \;\;\;\;k\in[0,n] \nonumber
\end{align}
which includes the state $\mathbf{x}$ for all $n$ frames and the visual
feature inverse depth $l$. The $k$th frame state $\mathbf{x}_k$ includes the
position $\mathbf{p}_{b_k}^{iw}$, velocity $\mathbf{v}_{b_k}^{iw}$, and
orientation in quaternions $\mathbf{q}_{b_k}^{iw}$ for robot $i$ in its world
frame, plus accelerometer bias $\mathbf{b}_{a_k}^{ib}$ and gyroscope bias
$\mathbf{b}_{w_k}^{ib}$ in its body frame.  $l_i$ is the inverse depth of the
$i$th feature among $m$ features from the visual observations over the sliding
window. If the sliding window includes all the camera frames since the start
of the mission, the optimization becomes a full smoothing estimation. Although
full smoothing offers the best accuracy, it is not scalable in reality, so we
use a key frame approach that discards similar frames while not losing
tracking. However, as the trajectory becomes longer, the size of state keeps
growing. For this reason, we fix the size of key frames and marginalize older
key frames into a prior factor in the optimization. The drift from visual
inertial odometry is still hard to avoid, and pose graph optimization with
loop closure becomes the only chance to correct the accumulating error.

Aiming at an accurate localization system and avoiding all rely on loop
closure, we design a SLAM system that uses a novel double layer sliding window
structure, with an implementation based on VINS-Mono~\cite{qin_vinsmono_2018}.

\subsection{Double Layer Tightly Coupled VIR Optimization}

We propose a double layer sliding window tightly coupled SLAM optimization by
considering following aspects: high accuracy relative pose estimation from
VIO, less accurate but absolute UWB measurements, and the computation cost for
these factors.

We design two sliding windows involving three kinds of variables, shown in
Equ.~\ref{equ:states}. A short window $\Psi$, the same as the classical visual
SLAM sliding window with size $n$ from Equ.\ref{equ:classicalState}. The
variables in this window include $\mathbf{x}_k$ (Equ.~\ref{equ:shortwindow})
and $[l_0, l_1, \dots, l_m]$, which have same meaning as the variables in
Equ.~\ref{equ:classicalState}. The novelty in our system lies in the addition
of another long sliding window $\Omega$, which carries state $\mathbf{w}_t$,
as shown in Equ. \ref{equ:longwindow}. The size of the long slide window is
$s$, which is much larger than $n$. The state in this window only contains the
robot position $\mathbf{p}_{b}^{iw}$ in robot world frame.
\begin{align}
\mathcal{X} &= [\mathbf{w}_0,\mathbf{w}_1,\dots,\mathbf{w}_s,\mathbf{x}_0,\mathbf{x}_1,\dots,\mathbf{x}_n,l_0, l_1, \dots, l_m] \label{equ:states}\\
\mathbf{x}_k &= [
\mathbf{p}_{b_k}^{iw}, \mathbf{v}_{b_k}^{iw}, \mathbf{q}_{b_k}^{iw}, \mathbf{b}_{a_k}^{ib}, \mathbf{b}_{w_k}^{ib}] \;\;\;\;\;\;\;\;\;\;\;\;\;k\in \Psi[0,n] \label{equ:shortwindow}\\
\mathbf{w}_t &= [
\mathbf{p}_{b_t}^{iw}]\;\;\;\;\;\;\;\;\;\;\;\;\;\;\;\;\;\;\;\;\;\;\;\;\;\;\;\;\;\;\;\;\;\;\;\;\;\;\;\;\; \;\; t\in \Omega[0,s] \label{equ:longwindow}
\end{align}
Fig.\ref{fig:factorgraph} shows the factor graph of our system, corresponding
to the scenario of Fig.~\ref{fig:problem}. The short window area in gray
includes the state from the camera and IMU measurements. The long window area in
orange contains the state from the UWB measurements. Following the state
definition, we formulate a full nonlinear optimization problem as:
\begin{figure}
\begin{center}
\includegraphics[width=0.45\textwidth,center]{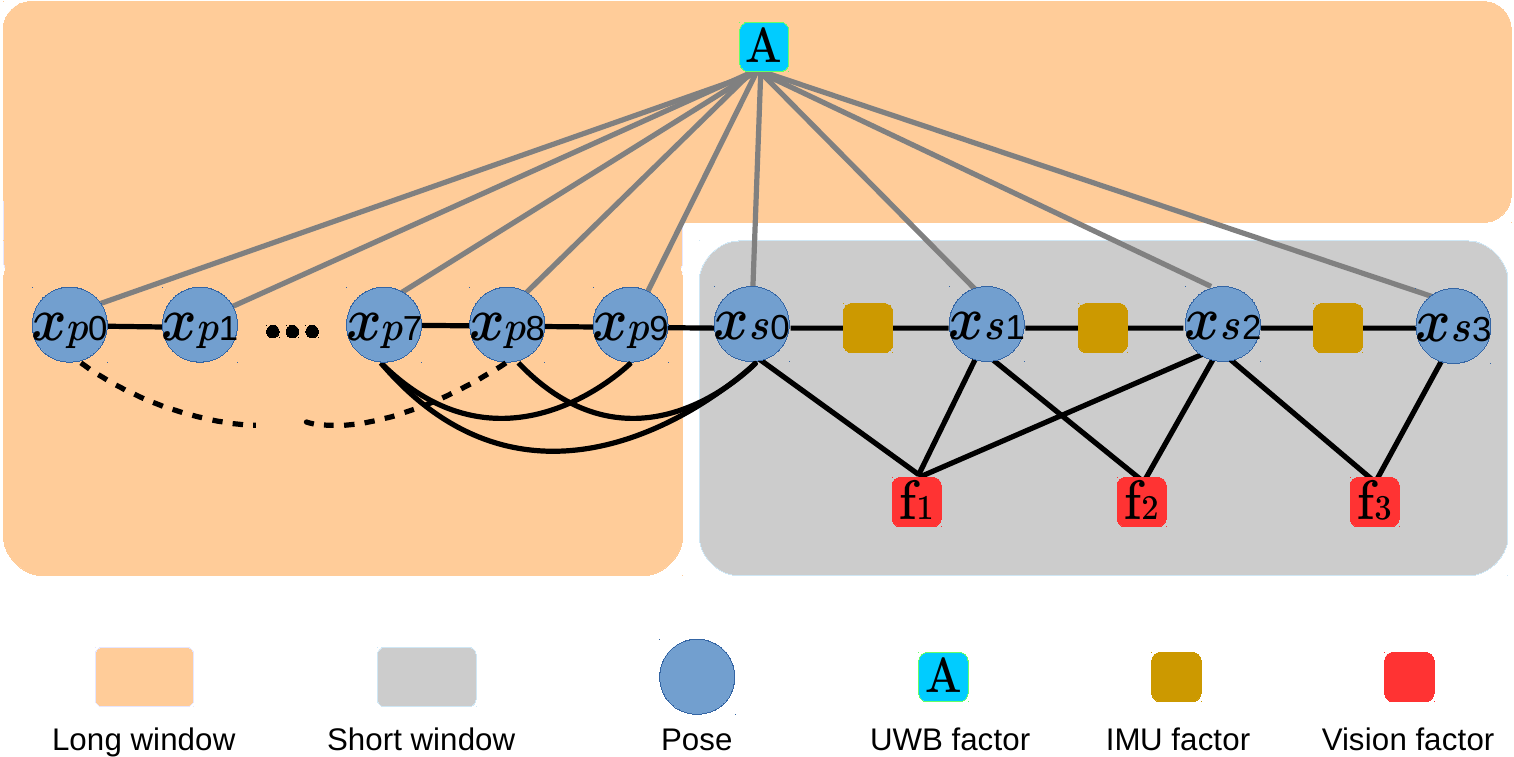}
\caption{Factor graph for the scenario in Fig. \ref{fig:problem}. The factor
  graph includes poses, UWB factor, IMU factors and visual factors. The edges
  are the error. We have a UWB factor for the poses with range
  measurements. Further, smooth edges (curves) are added between
  poses.}
\label{fig:factorgraph} 
\end{center}
\end{figure}

\begin{align}
\min _{\mathcal{X}}&
\Bigr\{
\underbrace{
\sum_{t \in\Omega}
{\rho(\|\mathbf{r}_{\mathcal{U}}(\hat{\mathbf{z}_t},\mathcal{X})\|}_{\mathbf{P}_{b_t}^{iw}})
}_{UWB\; factor}
+ 
\underbrace{
\sum_{k \in \Psi}\|\mathbf{r}_{\mathcal{B}}(\hat{\mathbf{z}}_{b_{k+1}}^{b_{k}}, \mathcal{X})\|_{\mathbf{P}_{b_{k+1}}^{b_{k}}}^{2}
}_{IMU\; factor}
\nonumber \\&
+
\underbrace{
\sum_{(l, j) \in \mathcal{C}} \rho(\|\mathbf{r}_{\mathcal{C}}(\hat{\mathbf{z}}_{l}^{c_{j}}, \mathcal{X})\|_{\mathbf{P}_{l}^{c_{j}}}^{2})
}_{Vision\; factor}
\Bigr\} \label{equ:optimization}
\end{align}


This nonlinear optimization problem considers three factors, corresponding to
the UWB factor, the IMU factor, and the vision factor. The UWB residuals are
calculated for the long window $\Omega$, while IMU and visual residuals for
the short window $\Psi$.  

\subsubsection{UWB Factor}

UWB localization uses different protocols: time of arrival (TOA), time
difference of arrival (TDoA), and two-way ranging (TWR). In our system, we use
TWR, which measures the distances between two transceivers by sending a packet
back and forth. Although the number of nodes supported is limited because of
the shared UWB communication medium, TWR can be used without device
synchronization, which makes the protocol widely used. Since we are not
considering hundreds of robot communicating simultaneously, we choose TWR to
avoid synchronization, which can be difficult to implement in distributed
multi-robot applications.
We model the ranging measurement of UWB modules as:
\[\label{uwb_model}
{\hat{d}} = {d} + {b}_{d} + {e}
\]
where ${\hat{d}}$ is the UWB measurement, ${d}$ is true distance, ${b}_{d}$ is
distance based bias, and ${e}$ is the error following a Gaussian distribution
${N}(0,\,\sigma^{2})$. Note that the ${b}_{d}$ can be calibrated with a Gaussian
process model beforehand by collecting data with ground truth. Then the model
can be simplified as:
\[\label{uwb_model1}
{\hat{d}} = {d} + {e}
\]
With this simplified UWB model, we define the UWB factor in
Equ.~\ref{equ:optimization} as:
\begin{align}
\mathbf{r}_{\mathcal{U}}(\hat{\mathbf{z}}_t,\mathcal{X}) =
\underbrace{
\gamma_r\cdot\rho( \|\mathbf{p}_{b_t}^{iw}-\mathbf{P}_A^{iw}\|-{\hat{d}}_t)}_{UWB\; ranging \; residual}
\nonumber\\
+
\underbrace{
\gamma_s\cdot(
\sum_{j\in(t,t+s]}
\{\mathbf{p}_{b_j}^{iw}-\mathbf{p}_{b_t}^{iw}\}-\widetilde{\mathbf{z}}_{b_{t}}^{b_{j}})
}_{Relative \; transformation \; residual}
\end{align}
The UWB factor above includes two residuals: a ranging measurement residual
and a virtual relative transformation measurement residual. $\gamma_r$ and
$\gamma_s$ are the weights for these two residuals. ${\hat{d}}_t$ are
the UWB ranging measurements, which are compared with the predicted ranging
from the robot body frame $t$ to the anchor $\mathbf{P}_A^{iw}$ in the world frame. To
avoid the ranging factor excessively affecting the optimizer and breaking the
relative pose property from visual-inertial estimation, we introduce a virtual
relative transformation measurement $\widetilde{\mathbf{z}}_{b_{t}}^{b_{j}}$
between frames $t$ and $j\in(t,t+s]$, which is extracted from the short sliding
window estimation result. We select $s=3$ in our experiments
to create two consecutive links between poses, as can be seen for example with $x_{p7}$ in
Fig.~\ref{fig:factorgraph}.  $\rho()$ is a pseudo-Huber loss function
defined as $\rho(q)=\delta^{2}(\sqrt{1+(q / \delta)^{2}}-1)$.



\subsubsection{IMU factor}

IMU measurements are critical for monocular visual odometry. As the frequency
of the IMU is usually higher than the camera image frame rate, the IMU
measurements are preintegrated between two consecutive image
frames~\cite{lupton_preintegration_2012}. By referring to the last body frame
motion, this technique avoid repeated IMU reintegration and reduces
computation during optimization. Forster et al.~\cite{forster_manifold_2017}
extend this approach to manifold structures of the rotation group
$\mathfrak So3$ for higher accuracy and robustness. We follow the same method
as~\cite{qin_vinsmono_2018} in quaternion form.

The preintegrated IMU measurements between two consecutive frames of $b_k$ and
$b_{k+1}$ referring to frame $b_k$ can be expressed as:
\begin{align*}
\boldsymbol{\alpha}_{b_{k+1}}^{b_{k}}&=\iint_{t \in\left[t_{k}, t_{k+1}\right]} \mathbf{R}_{t}^{b_{k}}\left(\hat{\mathbf{a}}_{t}-\mathbf{b}_{a_{t}}\right) d t^{2}\\
\boldsymbol{\beta}_{b_{k+1}}^{b_{k}}&=\int_{t \in\left[t_{k}, t_{k+1}\right]} \mathbf{R}_{t}^{b_{k}}\left(\hat{\mathbf{a}}_{t}-\mathbf{b}_{a_{t}}\right) d t \\
\boldsymbol{\gamma}_{b_{k+1}}^{b_{k}}&=\int_{t \in\left[t_{k}, t_{k+1}\right]} \frac{1}{2} \mathbf{\Omega}\left(\hat{\boldsymbol{\omega}}_{t}-\mathbf{b}_{w_{t}}\right) \gamma_{t}^{b_{k}} d t
\end{align*}
where
\[
\mathbf{\Omega}(\boldsymbol{\omega})=\left[\begin{array}{cc}
{-\lfloor\boldsymbol{\omega}\rfloor_{\times}} & {\boldsymbol{\omega}} \\
{-\boldsymbol{\omega}^{T}} & {0}
\end{array}\right],\lfloor\boldsymbol{\omega}\rfloor_{\times}=\left[\begin{array}{ccc}
{0} & {-\omega_{z}} & {\omega_{y}} \\
{\omega_{z}} & {0} & {-\omega_{x}} \\
{-\omega_{y}} & {\omega_{x}} & {0}
\end{array}\right]
\]
$\hat{\mathbf{a}}_t$ and $\hat{\boldsymbol{\omega}}_t$ are the accelerometer
and gyroscope measurement vectors, respectively. These three formulas
correspond to relative the motion changes of position, velocity and orientation
to the local body frame of $b_k$.

The IMU factor is the residual between predicted motion and the preintegrated
results referring to the body frame:
\[
\mathbf{r}_{\mathcal{B}}\left(\hat{\mathbf{z}}_{b_{k+1}}^{b_{k}}, \mathcal{X}\right)=\left[\begin{array}{c}{\delta \boldsymbol{\alpha}_{b_{k+1}}^{b_{k}}} \\ {\delta \boldsymbol{\beta}_{b_{k+1}}^{b_{k}}} \\ {\delta \boldsymbol{\theta}_{b_{k+1}}^{b_{k}}} \\ {\delta \mathbf{b}_{a}} \\ {\delta \mathbf{b}_{g}}\end{array}\right]
=\left[\begin{array}{c}
{
\mathbf{p}^{b_k} - \hat{\boldsymbol{\alpha}}_{b_{k+1}}^{b_{k}}}\\
\mathbf{v}^{b_k} - \hat{\boldsymbol{\beta}}_{b_{k+1}}^{b_{k}}\\
\boldsymbol{\theta}^{b_k} \otimes(\hat{\boldsymbol{\gamma}}_{b_{k+1}}^{b_{k}})^{-1}\\
{\mathbf{b}_w^{b_{k+1}}-\mathbf{b}_w^{b_{k}}}\\
{\mathbf{b}_a^{b_{k+1}}-\mathbf{b}_a^{b_{k}}}
\end{array} \right]
\]
where
$ \mathbf{p}^{b_k} =
{\mathbf{R}_{w}^{b_{k}}\left(\mathbf{p}_{b_{k+1}}^{w}-\mathbf{p}_{b_{k}}^{w}+\frac{1}{2}
    \mathbf{g}^{w} \Delta t_{k}^{2}-\mathbf{v}_{b_{k}}^{w} \Delta
    t_{k}\right)}$ ,
$ \mathbf{v}^{b_k}
=\mathbf{R}_{w}^{b_{k}}\left(\mathbf{v}_{b_{k+1}}^{w}+\mathbf{g}^{w} \Delta
  t_{k}-\mathbf{v}_{b_{k}}^{w}\right) $ and
$\boldsymbol{\theta}^{b_k} = \mathbf{q}^{b_{k}}_{w} \otimes
\mathbf{q}_{b_{k+1}}^{w} $ are the estimated position, velocity and
orientation referred to the local body frame of $b_k$
(see~\cite{qin_vinsmono_2018} for details).

\subsubsection{Vision Factor}

The vision factor consists of the reprojection error for the tracked
features. We use a strategy similar to~\cite{qin_vinsmono_2018}, comparing the
reprojection of all features in the current frame with their first
observations. We define the visual residual as
\begin{align*}
\mathbf{r}_{\mathcal{C}}\left(\hat{\mathbf{z}}_{l}^{c_{j}}, \mathcal{X}\right)&= \| \mathbf{u}_{l_k}^{c_j} - \pi(\mathbf{R}_{c_i}^{c_j},\mathbf{T}_{c_i}^{c_j},\mathbf{P}_{l_k}^{c_i}) \|
\end{align*}
where $\mathbf{u}_{l_k}^{c_j}$ is the coordinate of feature $l_k$ in the image
of the camera frame $j$, $\pi()$ is the projection that converts homogeneous
coordinates into image coordinates,
$\mathbf{R}_{c_i}^{c_j},\mathbf{T}_{c_i}^{c_j}$ represent the frame
transformations (rotation and translation) from the camera frame $i$ to $j$,
which are inferred from state poses, and $\mathbf{P}_{l_k}^{c_i}$ is the 3D
position of the ${k}$th feature in the first observation frame $i$. The vision factor
iterates through all the frames and all the tracked features in the estimated
state.

\subsubsection{Anchor position estimation}

In the above discussion, we assume the anchor coordinates are available for
the optimizer. As the anchor position is initially unknown, it needs to be
estimated. Therefore, the state vector to estimate at the start becomes
$ \mathcal{X} =
[\mathbf{P}_A^{iw},\mathbf{w}_0,\mathbf{w}_1,\dots,\mathbf{w}_s,\mathbf{x}_0,\mathbf{x}_1,\dots,\mathbf{x}_n,l_0,
l_1, \dots, l_m] $. The cost function and factors are kept unchanged. The
optimization result of $\mathbf{P}_A^{iw} $ after the initialization stage is
saved as a fixed value. We fixed the anchor position with two considerations:
a) The anchor is static in practice, and b) usually, the initialization phase
can be controlled and the robot moves in proximity of the anchor. Although the
distance measurement error is not correlated with the distance value, the
estimation of the anchor position depends on the distance. In other words, the
ratio of trajectory length in the initialization with the distance affects the
estimation. So we treat the initial estimate as fixed value.

\subsection{Distributed Collaborative SLAM}

Another significant benefit of our ranging-aided system is that, with a common
anchor, robots can directly estimate inter-robot transformations when they
rendezvous. Robots simply need to send their current position and anchor
position while ranging their neighbors. After receiving this
information twice, a robot can calculate the transformation matrix between itself
and the sender. Once the transformation matrix is correctly estimated, all information
received from neighbors can be correctly placed in the robot's frame.

Estimating the transformation between robots is a critical requirement for
multi-robot systems. We mark the transformation of coordinate systems from
robot $i$ to $j$ as $\mathbf{T}_i^j = [\mathbf{R}_i^j, \mathbf{t}_i^j]$, where
$\mathbf{R}_i^j$ is $3\times3$ matrix representing rotation and
$\mathbf{t}_i^j$ is $3\times1$ vector of translation. With the help of
accelerometers and gyroscopes, we can establish the direction of gravity and
define the same $z$ axis for all robots. VINS-Mono~\cite{qin_vinsmono_2018}
uses the same strategy: the $z$ axis is aligned with the opposite of gravity when
creating the coordinate system. Therefore, only the yaw angle $\theta$ and 3D
offset $\mathbf{t}_i^j$ between two coordinate systems need to be
estimated. The transformation $\mathbf{T}_i^j$ consists of:

\[
\mathbf{R}_i^j=
\left[{\begin{array}{ccc}
{cos(\theta)} & {-sin(\theta)} & {0} \\
{sin(\theta)} & {cos(\theta)} & {0} \\
{0} & {0} & {1}
\end{array}}
\right], 
\mathbf{t}_i^j=
\left[{\begin{array}{c}
t_x\\
t_y\\
t_z
\end{array}}
\right]
\]
Using the estimation of the anchor position for two robots, we can easily
get $t_z = ({\mathbf{P}_{A}^{jw}}- \mathbf{P}_{A}^{iw})_z$, which is the
projection of the difference vector on $z$ axis. This leaves three parameters
($\theta, t_x, t_z$) to be estimated.

\begin{figure}[h]
		\includegraphics[width=0.75\columnwidth,center]{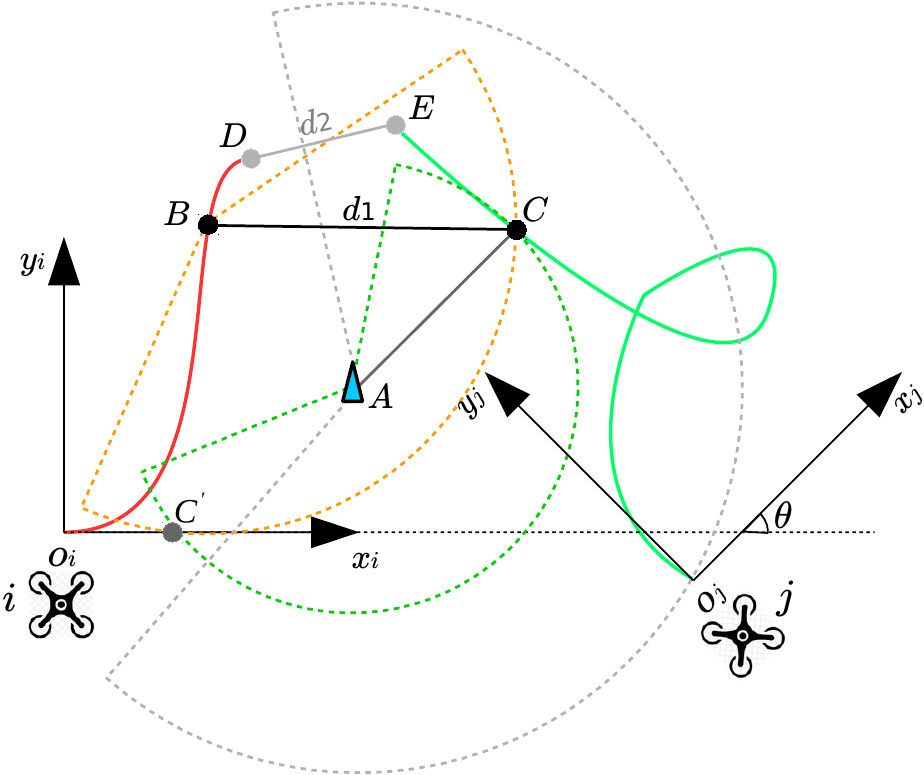}
		\caption{Transformation matrix estimation for multiple robots scenario.}
		\label{fig:uwbCoSlam} 
\end{figure}

Fig.\ref{fig:uwbCoSlam} shows how the remaining parameters are estimated.  Let
us assume two robots $i$ and $j$ moving independently in 2D for simplicity
(without loss of generality, as we already know the $z$ axis offset
$t_z$). Both robots have their own coordinate systems, $o_ix_iy_i$ and
$o_jx_jy_j$. They also have the anchor $A$ position after initialization, and
have their own trajectory tracked in their respective coordinate systems. When
the robot $i$ passes point $B$ and robot $j$ passes point $C$, they
enter in communication range and they can both obtain a reciprocal distance
measurement $d_1$. Simultaneously, they transmit their current
position and $A$'s position (which are in their own coordinate
system) to each other. 

We take the view of robot $i$ to illustrate our solution. When robot $i$
receives the position of the anchor and the current position of robot $j$
(point $C$) expressed in $j$'s frame, $\mathbf{P}_{b_A}^{jw}$ and
$\mathbf{P}_{b_C}^{jw}$, respectively. Robot $i$ then knows the origin of
$o_jx_jy_j$ must be on the gray circle around the anchor with radius $|Ao_j|$
(known from $\mathbf{P}_A^{jw}$). In addition, robot $i$ calculates $|AC|$ from
$\mathbf{P}_A^{jw}$ and $\mathbf{P}_{b_C}^{jw}$ to find robot $j$'s position $C$ must lie on a green circle around A with radius
$|AC|$. As we know, the distance between point $B$ and $C$ is the range
measurement $d_1$, and point C also must lie on the orange circle around its
current position $B$ with radius $d_1$.

Therefore, the current position lies at one of the intersections between the
green and orange circles, $C$ and $C'$. To find which intersection if the
correct one, we take an additional measurement $d_2$ between two following points
$D$ and $E$. Following the same procedure, we can find the candidates
of $E$. For clarity, we did not draw these circles for $D$ and $E$. By
comparing the relation between $E$, $E'$, $C$ and $C'$ with the true motion
from $C$ to $E$, we can find $C$ and determine the transformation
$\mathbf{T}_j^i$. This is the same as solving:
\begin{align}
\left\{ 
\begin{array}{rc}
\mathbf{P}_{b_A}^{jw} =& \mathbf{T}_i^j \cdot \mathbf{P}_{b_A}^{iw} \\
d_1 =& \|\mathbf{P}_{b_B}^{jw} - \mathbf{T}_i^j \cdot \mathbf{P}_{b_C}^{iw}\|\\
d_2 =& \|\mathbf{P}_{b_D}^{jw} - \mathbf{T}_i^j \cdot  \mathbf{P}_{b_E}^{iw}\|
\end{array}
\right.
\end{align}
where  $\mathbf{P}_{b_Q}^{rw}$ represent the 3D position vector of points $Q$ in world coordinate frame of robot $r$.

\section{EXPERIMENTS}

We validate our algorithms using public datasets and real hardware
experiments. We choose
VINS-Mono\footnote{https://github.com/HKUST-Aerial-Robotics/VINS-Mono} and
compare with it using the TUM evaluation
tool\footnote{https://vision.in.tum.de/data/datasets/rgbd-dataset/tools}.  We
compute the absolute trajectory error (ATE) when ground truth is
available. For our large area experiments we do not have ground truth, and we
start and end the experiment in the same location and then calculate the
start-to-end error.

\subsection{EuRoC Dataset Experiments}

We test our single robot tracking system on the EuRoC~\cite{burri_euroc_2016}
dataset. We simulate the UWB ranging measurements from ground truth
data. The static anchor is assumed in the origin of the frame created during
robot initialization. We add Gaussian white noise $\mathcal{N}(0,\,0.05)$ to model the
error of our UWB sensor. The sliding window size of VINS-Mono is set to
10. For our system, we tested with a short window size of 10 and a long window
size of 100.

\begin{figure}[tbp]
\begin{center}
\includegraphics[width=0.85\columnwidth,center]{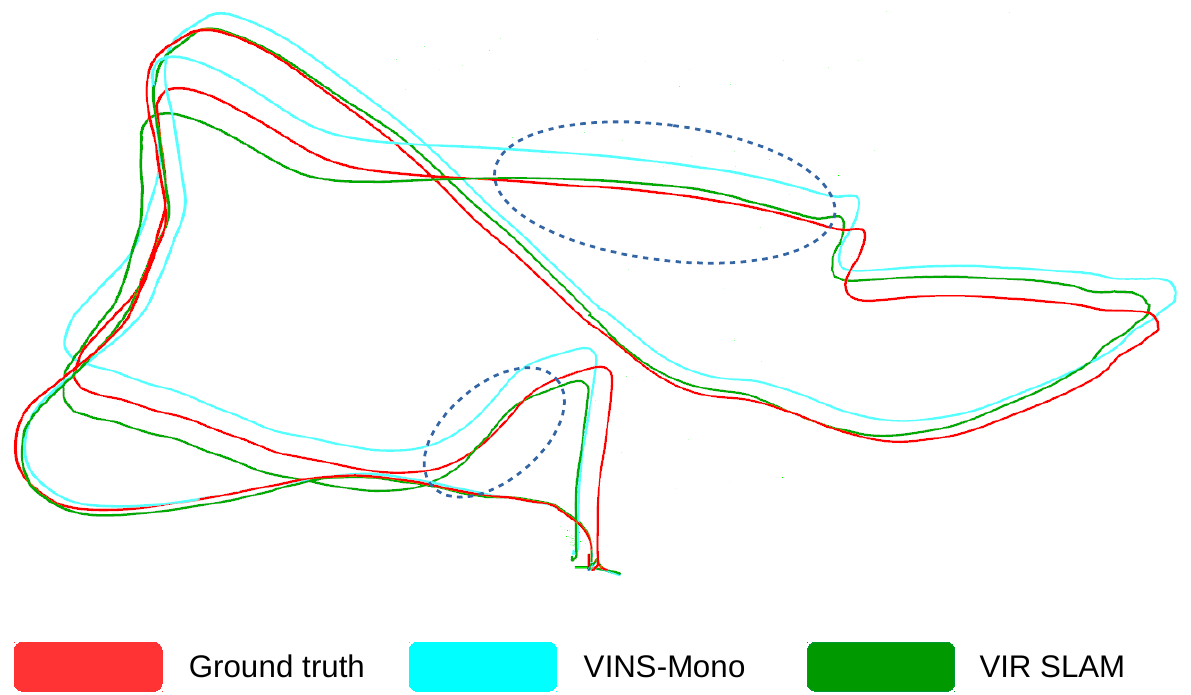}
\caption{VIR results in EuRoC MH05 Dataset. ATE error of our method is 0.291m, compared to VINS-Mono MONO (0.388m). We set our method is closer to the ground truth in most sections, which testify the capability to correct the drift error.}
\label{fig:euroc05} 
\end{center}
\end{figure}

Table~\ref{table:ATE} shows our VIR-SLAM outperform VINS-Mono in terms of
ATE. As an example, we show the trajectory of the
comparison for the EuRoC MH-05 sequence in~Fig.\ref{fig:euroc05}. The green
trajectory of our method is closer to the red ground truth. From the
ellipse indicator, we can see our estimation can correct the error even when drifting.

\begin{table}[tbp]
\caption{Error comparison between VINS-Mono and our VIR SLAM.}
\begin{center}
\begin{tabular}{|c||c|c|c|}
\hline
ATE Error (m)  & VINS-Mono & \textbf{VIR} SLAM & Improvement\\
\hline
\textbf EuRoC MH\_01 &  0.186 & 0.178 & 4.33 \% \\
\textbf EuRoC MH\_02 &  0.240 &  0.188 & 21.76 \%  \\
\textbf EuRoC MH\_03 &  0.271 &   0.260 &4.04 \% \\
\textbf EuRoC MH\_04 &  0.402 &   0.366 & 9.0 \% \\
\textbf EuRoC MH\_05 &  0.388 &   0.291 &24.84 \% \\
\textbf Lab Seq1 &  0.266 &   0.193 & 27.81 \% \\
\textbf Lab Seq2 &  0.195 &   0.169 &15.08 \% \\
\hline
\end{tabular}
\end{center}
\label{table:ATE}
\end{table}

\subsection{Single Robot experiments}

\begin{figure}[tbp]

		\includegraphics[width=0.8\columnwidth,center]{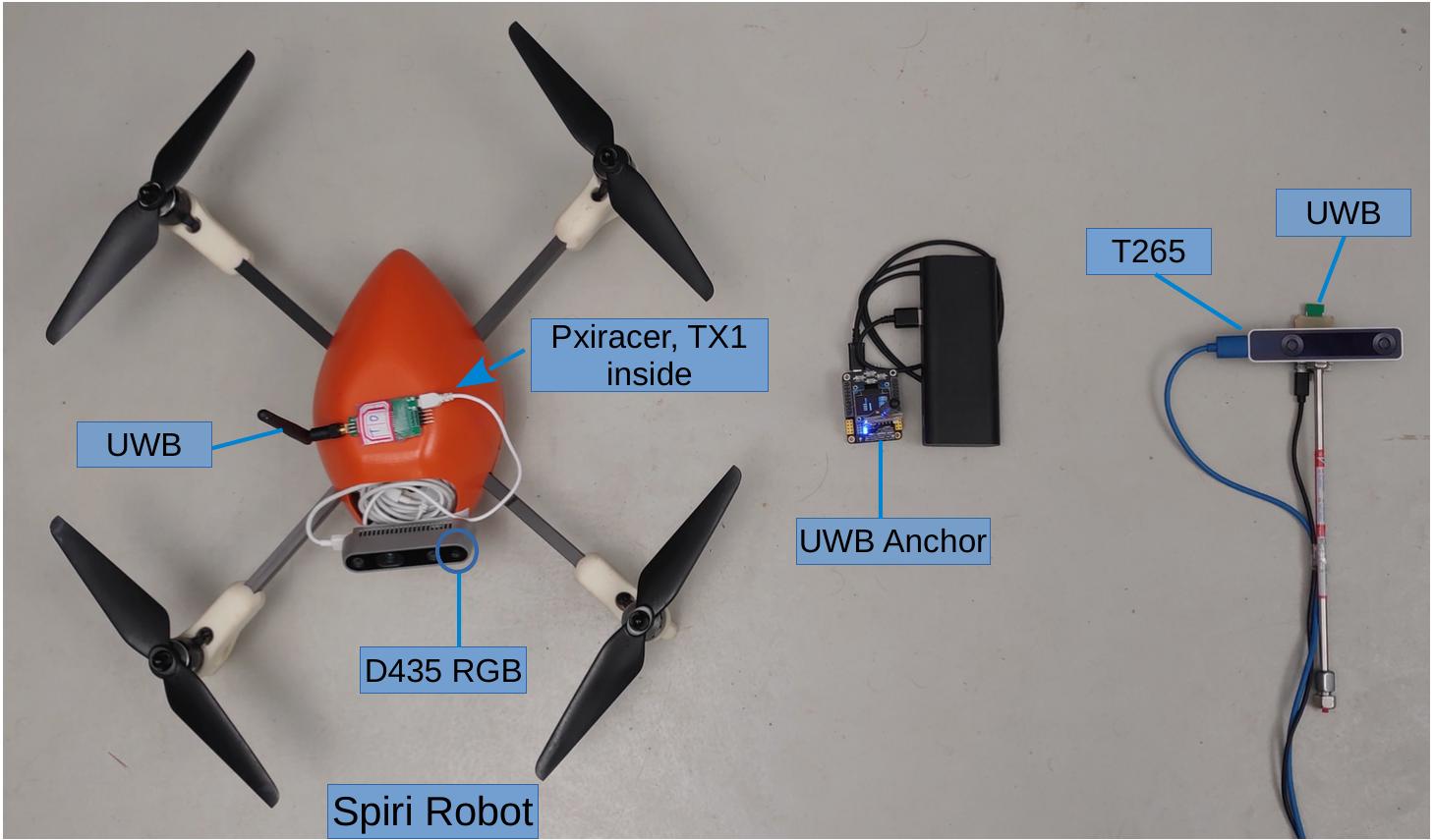}
		\caption{For single robot experiments, we use the Spiri robot
                  (w/ Pixracer and NVIDIA TX1). We add a Realsense D435 camera
                  (although we only use RGB) and UWB sensor module. The IMU
                  measurements are obtained from the Pixracer. The
                  static is in the middle, while the Realsense T265 on the right
                  simulates a second robot in our multi-robot experiment.}
		\label{fig:robot} 
\end{figure}

We also test our system with a Spiri robot with real UWB sensors, shown in
Fig.~\ref{fig:robot}. The system has a D435 Realsense camera but we use it as
a monocular camera. The IMU measurements come from the Pixracer flight
controller, and the robot carries an Nvidia TX1 as an on-board computer. We
test the system in our lab with an OptiTrack motion capture system as ground
truth.

\begin{figure}[tbp]
\includegraphics[width=0.9\columnwidth,center]{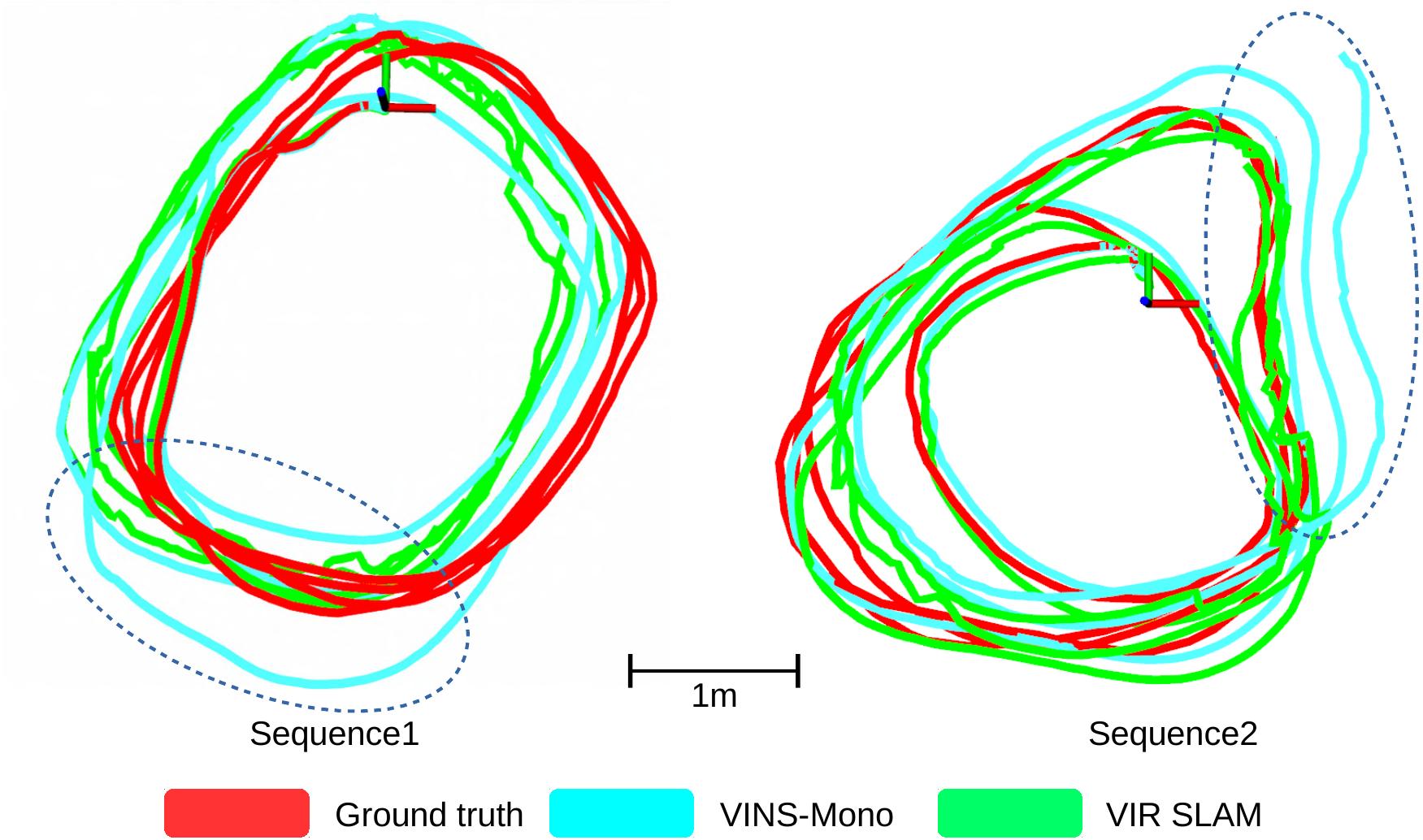}
\caption{VIR indoor experiments with OptiTrack ground truth. The left and
  right figures are two sequences. All trajectories start at the origin and
  are not aligned. Our estimator does not present excessive drift.}
\label{fig:indoorExp} 
\end{figure}

We manually control the robot and collect two sequences: the results are
listed in Table~\ref{table:ATE}. To clearly see the difference, we show the
trajectories from same origin. This is different from aligning the two
trajectory to compute the ATE. As Fig.~\ref{fig:indoorExp} shows, our
estimator has less accumulated error than VINS-Mono.

\begin{figure}[tpb]
\includegraphics[trim=0cm 0cm 1.0cm 1.7cm,clip, width=0.65\columnwidth,center]{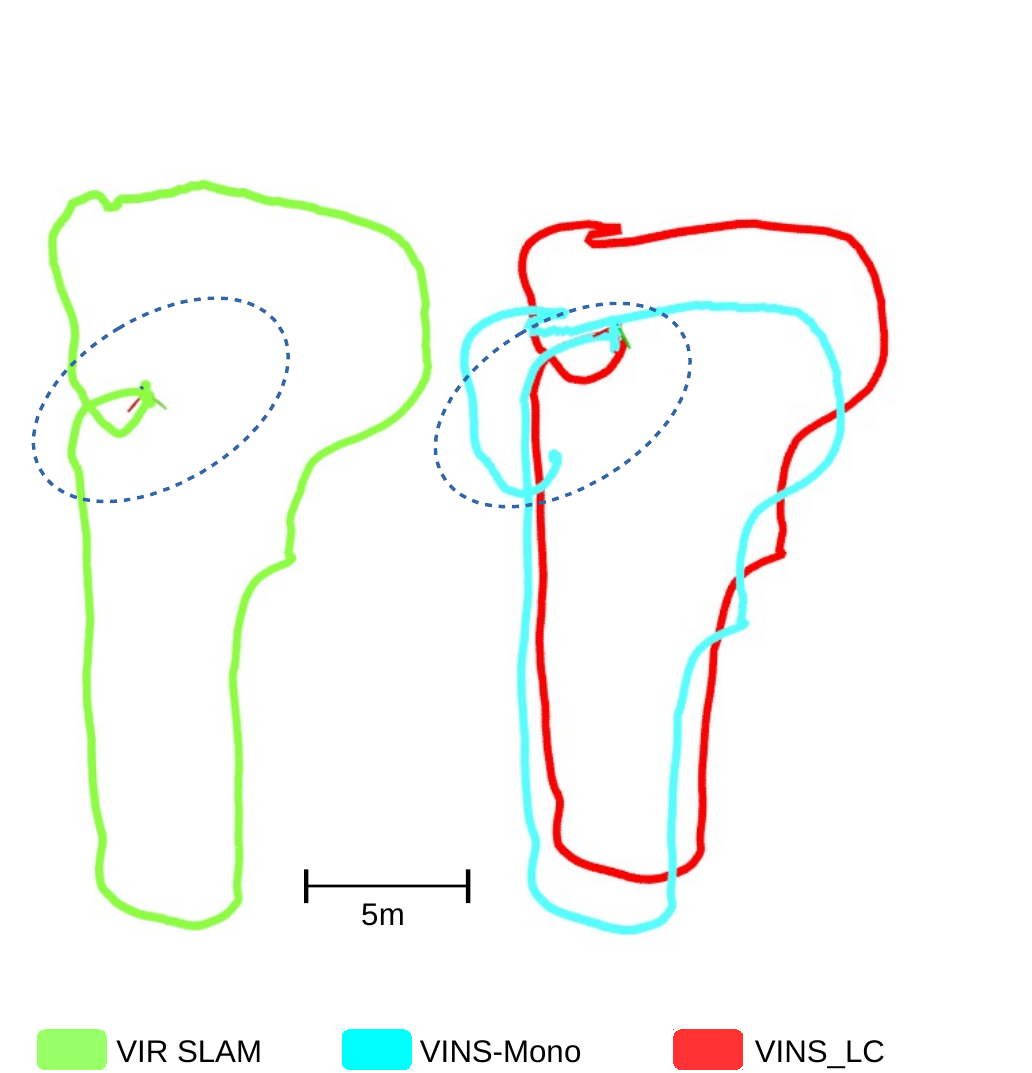}
\caption{VIR results in the atrium, which is very visually challenging. The
  robot starts and ends at the same point and we compare the start-to-end
  error. As the figure shows, the start and end point of VIR-SLAM are
  overlapped. Although the VINS-Mono with loop closure (VINS\_LC) can close
  the loop, it does not eliminate the accumulated error from the VIO.}
\label{fig:atriumExp} 
\end{figure}

We also test our system performance in large atrium. The environment in the
atrium is quite challenging: featureless walls, low light conditions, glass
walls with reflections, etc.  We move the robot around the atrium and make
sure the robot ends at the same point as it starts. Then we compare the
start-to-end error. As Fig.~\ref{fig:atriumExp} and
Table~\ref{table:Start-to-end Error} shows, the VINS-Mono estimation has
significant drift, more than 5.4m. Although the loop closure version can
correct the drift as the loop closure is detected, there is still a clear
error in the trajectory. However, our VIR SLAM works correctly without any
loop closure. The start and end point have a small translation error (0.148m)
in 2D, but our system introduces a bigger accumulated error on the $z$
axis. We believe this is due to the anchor being close to the horizontal plane
of the robot, and the small difference in measurement does not help correct
the $z$ coordinate. We manually checked the UWB ranging information along the
top and bottom edges and we can confirm that our estimation is at the right
position on the boundaries.
\begin{table}[htbp]
	\caption{Start-to-end Error Comparison}
		\begin{tabular}{|c||c|c|c|}
			\hline
			   & End point & 2D Error(m) & 3D Error(m)\\
			\hline
			\textbf VINS-Mono &  (4.29,	3.35, 0.52) & 5.439 & 5.465 \\
			\textbf VINS-LC SLAM  & (0.18, 0.43, -0.38) &  0.464 & \textbf{0.602}\\
			\textbf  VIR-SLAM &  (-0.04, 0.14, 0.84) & \textbf{0.148} &0.853\\
			\hline
		\end{tabular}
	\label{table:Start-to-end Error}
\end{table}

\subsection{Multi-Robot Experiments}

We also test our multi-robot technique, as shown in
Fig.~\ref{fig:multirobotExp}. We manually and independently control two robots
to make a trajectory similar to the "MIST", the name of our lab. 
We place one static UWB anchor in the environment. The first
robot is imply a Realsense T265 and an UWB module, show in
Fig.~\ref{fig:robot} (right). We move it to form an "M" shape. The other
robot, Spiri, moves along a "IST" and it controlled independently. Their
trajectories in each robot's frame are shown at the bottom of
Fig.~\ref{fig:multirobotExp}. With simply two inter-robot measurements (with data exchange), robot
1 can estimate the transformation of robot 2 and map robot 2's
trajectory in its own frame as shown at the top of Fig.~\ref{fig:multirobotExp}.

\section{CONCLUSIONS AND DISCUSSIONS}

In this paper, we propose VIR-SLAM, a novel SLAM paradigm combining vision,
inertial, and UWB sensors. By arbitrarily setting up a static UWB anchor in
the environment, robots can have drift-free state estimation and collaboration
on SLAM. Our solution combines the accurate relative pose estimation from VIO
and enhances it using ranging to correct the accumulated error. As our
experiments show, the introduced static anchor can help correct the drift
effectively to improve localization accuracy. We also show an example with two
independent robots mapping each other's
trajectory to their own frame after obtaining two range measurements. This
technique allows the robots to find the inter-robot transformation, which is
extremely useful for multi-robot SLAM.

\bibliographystyle{IEEEtran}

\bibliography{IEEEabrv,uvi_slam}

\begin{thebibliography}{10}
\providecommand{\url}[1]{#1}
\csname url@samestyle\endcsname
\providecommand{\newblock}{\relax}
\providecommand{\bibinfo}[2]{#2}
\providecommand{\BIBentrySTDinterwordspacing}{\spaceskip=0pt\relax}
\providecommand{\BIBentryALTinterwordstretchfactor}{4}
\providecommand{\BIBentryALTinterwordspacing}{\spaceskip=\fontdimen2\font plus
\BIBentryALTinterwordstretchfactor\fontdimen3\font minus
  \fontdimen4\font\relax}
\providecommand{\BIBforeignlanguage}[2]{{%
\expandafter\ifx\csname l@#1\endcsname\relax
\typeout{** WARNING: IEEEtran.bst: No hyphenation pattern has been}%
\typeout{** loaded for the language `#1'. Using the pattern for}%
\typeout{** the default language instead.}%
\else
\language=\csname l@#1\endcsname
\fi
#2}}
\providecommand{\BIBdecl}{\relax}
\BIBdecl

\bibitem{delmerico_benchmark_2018}
J.~Delmerico and D.~Scaramuzza, ``A benchmark comparison of monocular
  visual-inertial odometry algorithms for flying robots,'' in \emph{2018 {IEEE}
  International Conference on Robotics and Automation ({ICRA})}.\hskip 1em plus
  0.5em minus 0.4em\relax {IEEE}, pp. 2502--2509.

\bibitem{lupton_preintegration_2012}
T.~Lupton and S.~Sukkarieh, ``Preintegration\_visual-{Inertial}-{Aided}
  {Navigation} for {High}-{Dynamic} {Motion} in {Built} {Environments}
  {Without} {Initial} {Conditions},'' \emph{IEEE Transactions on Robotics},
  vol.~28, no.~1, pp. 61--76, Feb. 2012.

\bibitem{forster_manifold_2017}
C.~Forster, L.~Carlone, F.~Dellaert, and D.~Scaramuzza,
  ``\BIBforeignlanguage{en}{On-{Manifold} {Preintegration} for {Real}-{Time}
  {Visual}--{Inertial} {Odometry}},'' \emph{\BIBforeignlanguage{en}{IEEE
  Transactions on Robotics}}, vol.~33, no.~1, pp. 1--21, Feb. 2017.

\bibitem{qin_vinsmono_2018}
T.~Qin, P.~Li, and S.~Shen, ``{VINS}-{Mono}: {A} {Robust} and {Versatile}
  {Monocular} {Visual}-{Inertial} {State} {Estimator},'' \emph{IEEE
  Transactions on Robotics}, vol.~34, no.~4, pp. 1004--1020, Aug. 2018.

\bibitem{noauthor_loco_nodate}
\BIBentryALTinterwordspacing
``Loco {Positioning} system {\textbar} {Bitcraze}.'' [Online]. Available:
  \url{https://www.bitcraze.io/loco-pos-system/}
\BIBentrySTDinterwordspacing

\bibitem{saeedi_multi_2016}
S.~Saeedi, M.~Trentini, M.~Seto, and H.~Li, ``multi robot slam review
  {Multiple}-{Robot} {Simultaneous} {Localization} and {Mapping}: {A}
  {Review},'' \emph{Journal of Field Robotics}, vol.~33, no.~1, pp. 3--46,
  2016.

\bibitem{cadena_slam_2016}
C.~Cadena, L.~Carlone, H.~Carrillo, Y.~Latif, D.~Scaramuzza, J.~Neira, I.~Reid,
  and J.~J. Leonard, ``{Past}, present, and future of simultaneous localization
  and mapping: {Toward} the robust-perception age,'' \emph{IEEE Transactions on
  Robotics}, vol.~32, no.~6, pp. 1309--1332, 2016.

\bibitem{engel_direct_2018}
J.~Engel, V.~Koltun, and D.~Cremers, ``Direct {Sparse} {Odometry},'' \emph{IEEE
  Transactions on Pattern Analysis and Machine Intelligence}, vol.~40, no.~3,
  pp. 611--625, Mar. 2018.

\bibitem{prorok_accurate_2014}
A.~Prorok and A.~Martinoli, ``Accurate indoor localization with ultra-wideband
  using spatial models and collaboration,'' \emph{The International Journal of
  Robotics Research}, vol.~33, no.~4, pp. 547--568, 2014, 05.

\bibitem{mueller_fusing_2015}
M.~W. Mueller, M.~Hamer, and R.~D'Andrea, ``\BIBforeignlanguage{en}{Fusing
  ultra-wideband range measurements with accelerometers and rate gyroscopes for
  quadrocopter state estimation},'' in \emph{\BIBforeignlanguage{en}{2015
  {IEEE} {International} {Conference} on {Robotics} and {Automation}
  ({ICRA})}}.\hskip 1em plus 0.5em minus 0.4em\relax Seattle, WA, USA: IEEE,
  May 2015, pp. 1730--1736.

\bibitem{fang_graph_2018}
X.~Fang, C.~Wang, T.-M. Nguyen, and L.~Xie, ``Graph {Optimization} {Approach}
  to {Localization} with {Range} {Measurements},'' \emph{arXiv:1802.10276
  [cs]}, Feb. 2018, arXiv: 1802.10276.

\bibitem{tiemann_enhanced_2018}
J.~Tiemann, A.~Ramsey, and C.~Wietfeld, ``\BIBforeignlanguage{en}{Enhanced
  {UAV} {Indoor} {Navigation} through {SLAM}-{Augmented} {UWB}
  {Localization}},'' in \emph{\BIBforeignlanguage{en}{2018 {IEEE}
  {International} {Conference} on {Communications} {Workshops} ({ICC}
  {Workshops})}}.\hskip 1em plus 0.5em minus 0.4em\relax Kansas City, MO: IEEE,
  May 2018, pp. 1--6.

\bibitem{wang_ultra-wideband_2017}
C.~Wang, H.~Zhang, T.-M. Nguyen, and L.~Xie, ``Ultra-wideband aided fast
  localization and mapping system,'' in \emph{2017 IEEE/RSJ International
  Conference on Intelligent Robots and Systems (IROS)}.\hskip 1em plus 0.5em
  minus 0.4em\relax IEEE, 2017, pp. 1602--1609.

\bibitem{shi_visual-uwb_nodate}
Q.~Shi, X.~Cui, W.~Li, Y.~Xia, and M.~Lu,
  ``\BIBforeignlanguage{en}{Visual-{UWB} {Navigation} {System} for {Unknown}
  {Environments}},'' p.~11.

\bibitem{burri_euroc_2016}
M.~Burri, J.~Nikolic, P.~Gohl, T.~Schneider, J.~Rehder, S.~Omari, M.~W.
  Achtelik, and R.~Siegwart, ``\BIBforeignlanguage{en}{The {EuRoC} micro aerial
  vehicle datasets},'' \emph{\BIBforeignlanguage{en}{The International Journal
  of Robotics Research}}, vol.~35, no.~10, pp. 1157--1163, Sep. 2016.

\bibitem{saeedi_multiple-robot_2016}
S.~Saeedi, M.~Trentini, M.~Seto, and H.~Li, ``Multiple-{Robot} {Simultaneous}
  {Localization} and {Mapping}: {A} {Review},'' \emph{Journal of Field
  Robotics}, vol.~33, no.~1, pp. 3--46, 2016.

\bibitem{schmuck_ccm-slam_2019}
P.~Schmuck and M.~Chli, ``\BIBforeignlanguage{en}{{CCM}-{SLAM}: {Robust} and
  efficient centralized collaborative monocular simultaneous localization and
  mapping for robotic teams},'' \emph{\BIBforeignlanguage{en}{Journal of Field
  Robotics}}, vol.~36, no.~4, pp. 763--781, 2019.

\bibitem{choudhary_distributed_2017}
S.~Choudhary, L.~Carlone, C.~Nieto, J.~Rogers, H.~I. Christensen, and
  F.~Dellaert, ``Distributed {Mapping} with {Privacy} and {Communication}
  {Constraints}: {Lightweight} {Algorithms} and {Object}-based {Models},''
  \emph{arXiv:1702.03435 [cs]}, Feb. 2017, arXiv: 1702.03435.

\bibitem{mangelson_pairwise_2018}
J.~G. Mangelson, D.~Dominic, R.~M. Eustice, and R.~Vasudevan, ``Pairwise
  consistent measurement set maximization for robust multi-robot map merging,''
  in \emph{2018 {IEEE} {International} {Conference} on {Robotics} and
  {Automation} ({ICRA})}.\hskip 1em plus 0.5em minus 0.4em\relax IEEE, 2018,
  pp. 2916--2923.

\bibitem{trawny_3d_nodate}
N.~Trawny, X.~S. Zhou, and S.~I. Roumeliotis, ``3d relative pose estimation
  from six distances.'' 2009.

\bibitem{martel_unique_2019}
F.~M. Martel, J.~Sidorenko, C.~Bodensteiner, M.~Arens, and U.~Hugentobler,
  ``\BIBforeignlanguage{en}{Unique 4-{DOF} {Relative} {Pose} {Estimation} with
  {Six} {Distances} for {UWB}/{V}-{SLAM}-{Based} {Devices}},''
  \emph{\BIBforeignlanguage{en}{Sensors}}, vol.~19, no.~20, p. 4366, Oct. 2019.

\end{thebibliography}

\end{document}